\documentclass[conference, hidelinks]{IEEEtran}
\IEEEoverridecommandlockouts
\usepackage{cite}
\usepackage{algorithm}
\usepackage[noend]{algpseudocode}
\usepackage{graphicx}
\usepackage{xcolor}
\usepackage{multirow}
\usepackage{booktabs}
\usepackage{color}
\usepackage{enumerate}
\usepackage[inline]{enumitem}
\usepackage{listings}
\usepackage{multirow}
\usepackage{soul}
\usepackage{syntax}
\usepackage[referable]{threeparttablex}
\usepackage{xspace}
\usepackage[skip=2pt]{caption}
\usepackage{subcaption}

\usepackage{hyperref}
\hypersetup{colorlinks=false}

\usepackage{mathptmx}
\usepackage{newtxmath}

\DeclareMathOperator*{\argmin}{arg\,min}

\usepackage{etoolbox}
\usepackage{booktabs}
\usepackage{multirow}
\usepackage{tabularx}

\makeatother
\begin{document}

\newcommand{\name}{COSMA\xspace}

\newcommand{\yi}[1]{\textcolor{orange}{[YL: #1]}}
\newcommand{\sharad}[1]{\textcolor{red}{[SM: #1]}}
\newcommand{\aarti}[1]{\textcolor{blue}{[AG: #1]}}

\title{Combined Scheduling, Memory Allocation and Tensor Replacement for Minimizing Off-Chip Data Accesses of DNN Accelerators

\thanks{This work was supported in part by the Applications Driving Architectures (ADA) Research Center, a JUMP Center co-sponsored by SRC and DARPA. This work is also supported by Qualcomm Innovation Fellowship}

}


\author{\IEEEauthorblockN{Yi Li, Aarti Gupta, Sharad Malik}
\IEEEauthorblockA{Princeton University, Princeton, New Jersey, USA, \\
yi-li@princeton.edu, aartig@cs.princeton.edu, sharad@princeton.edu} 
}

\maketitle

\begin{abstract}
Specialized hardware accelerators have been extensively used for Deep Neural Networks (DNNs) to provide power/performance benefits. These accelerators contain specialized hardware that supports DNN operators, and scratchpad memory for storing the tensor operands. Often, the size of the scratchpad is insufficient to store all the tensors needed for the computation, and additional data accesses are needed to move tensors back and forth from host memory during the computation with significant power/performance overhead. The volume of these additional data accesses depends on the operator schedule, and memory allocation (specific locations selected for the tensors in the scratchpad).
We propose an optimization framework, named \name, for mapping DNNs to an accelerator that finds the optimal operator schedule, memory allocation and tensor replacement that minimizes the additional data accesses.
\name provides an Integer Linear Programming (ILP) formulation to generate the optimal solution for mapping a DNN to the accelerator for a given scratchpad size.
We demonstrate that, using an off-the-shelf ILP solver, \name obtains the optimal solution in seconds for a wide-range of state-of-the-art DNNs for different applications. Further, it out-performs existing methods by reducing on average 84\% of the non-compulsory data accesses.
We further propose a divide-and-conquer heuristic to scale up to certain complex DNNs generated by Neural Architecture Search, and this heuristic solution reduces on average 85\% data accesses compared with other works.
\end{abstract}

\section{Introduction}
Specialized hardware accelerators are increasingly utilized to enhance the power/performance efficiency of Deep Neural Networks (DNNs), particularly for DNN inference on power-constrained edge devices.
While emerging DNN accelerators can handle complex operators, their efficacy is often constrained by the available on-chip scratchpad memory.
If the on-chip scratchpad lacks the necessary capacity to store the tensors required for subsequent operator execution, tensors must be temporarily transferred to the host memory and later retrieved, which results in additional off-chip data accesses. 
As highlighted by previous studies~\cite{MarkEnergy}, off-chip data accesses consume considerably more power than accelerator computations, thereby compromising power efficiency and increasing execution latency.
Consequently, it is crucial for the compiler to minimize these additional data acesses when deploying DNNs on accelerators.

The amount of off-chip data accesses due to limited on-chip memory is determined by three factors: \textit{scheduling of DNN operators}, \textit{memory allocation of tensors}, and \textit{tensor replacement due to limited memory}.
Existing deep learning compiler frameworks, e.g., PyTorch~\cite{pytorch}, TensorFlow (TF)~\cite{tensorflow} and TVM~\cite{chen2018tvm}, often lack awareness of memory placement of tensors when mapping DNN operators to hardware. 
Typically, these frameworks schedule operators based on data dependencies or user-defined sequences, and they lean on traditional dynamic memory allocators for tensor memory management.
This approach can lead to memory fragmentation and sub-optimal memory utilization.
Moreover, these compiler frameworks offer limited support for custom hardware accelerators, where intensive manual effort is required to explicitly manage scheduling of operators and tensor memory management.
While TVM can identify efficient schedules for mapping individual operators by auto-tuning~\cite{chen2018tvm}, it does not extend this capability to scheduling and memory management for mapping entire, and even partial, DNN graphs to target accelerators, which is the focus of this paper.


Prior works~\cite{serenity, hmcos, telamalloc, model} address different aspects of the operator scheduling and tensor memory management problem, but none have tried to solve it in a unified way.
To address this gap we introduce \name which, to the best of our knowledge, is the first to collectively optimize the three dimensions of  operator scheduling, memory allocation and tensor replacement to minimize the amount of off-chip data accesses. 
In doing so, we make the following contributions:
\begin{itemize}[leftmargin=*]
    \item We propose an ILP formulation that combines operator scheduling, memory allocation and tensor replacement for mapping a DNN to a target accelerator with the objective of minimizing off-chip data accesses.
    \item We demonstrate that an off-the-shelf ILP solver can provide optimal solutions in seconds for a wide-range of state-of-the-art DNNs for different applications.
    \item We demonstrate the versatility of \name's formulation by modifying its objective function for different goals.
    \item We develop a divide-and-conquer heuristic to scale up the solution techniques to handle complex DNN models that were beyond the scale of previous techniques.
    \item We conduct an extensive evaluation that demonstrates the efficacy of our approach and its advantages over existing solutions over a range of popular DNNs.
\end{itemize}


\section{Background and Related Work}
\begin{figure}[t]
    \centering
    \captionsetup{font=small}
    \includegraphics[width=0.48\textwidth]{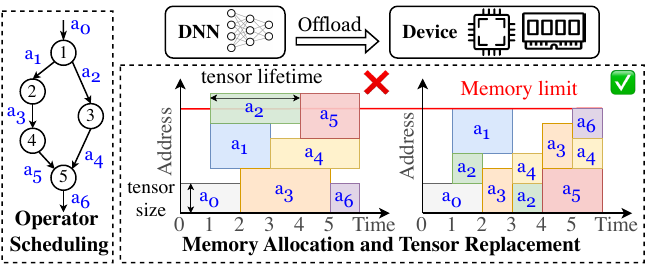}
    \caption{
    Example of mapping a DNN to an accelerator.
    Operator scheduling determines the node order in the DNN graph (number labeled on the node.)
    Tensors (labeled as $a_{i}$ on the edges) need to be allocated in the accelerator's on-chip memory.
    Due to the tight memory budget, the memory allocation scheme shown on the left is not valid.
    One valid memory allocation shown on the right has tensors $a_{2}, a_{3}, a_{4}$ spilled and later retrieved to keep total memory usage within the limit.
    }
    \label{fig:overall-prob}
\end{figure}

DNNs can be represented as dataflow graphs where nodes represent operators (e.g., convolution), and directed edges represent the tensor (i.e., data) produced by its source node and consumed by its sink nodes.
As shown in Fig.\ref{fig:overall-prob}, offloading a DNN to a target accelerator requires determining operator scheduling (i.e., order in which the graph nodes will be computed by the accelerator), memory allocation of the input/output tensors of the computation (i.e., location in the accelerator memory where each tensor will reside) and also tensor replacement if the remaining available memory space is not enough. 


\begin{table}[t]
\captionsetup{font=small}
\caption{Comparison With Related Work}
\label{tab:my-table}
\centering
\setlength\tabcolsep{1pt}
\begin{tabular}{@{}l|l|l@{}}
\toprule
\multicolumn{1}{c|}{Work} &
  \multicolumn{1}{c|}{Approach} &
  \multicolumn{1}{c}{Comments} \\ \midrule
\begin{tabular}[c]{@{}l@{}}PyTorch\cite{pytorch}\\ TF\cite{tensorflow}\\ TVM\cite{chen2018tvm}\end{tabular} &
  \begin{tabular}[c]{@{}l@{}}Memory-unaware schedules,\\ Heuristic-based memory \\ allocation \& replacement\end{tabular} &
  \begin{tabular}[c]{@{}l@{}}Limited support for accelerators,\\ Manual effort required\end{tabular} \\ \midrule
\begin{tabular}[c]{@{}l@{}}Serenity\cite{serenity}\\ HMCOS\cite{hmcos}\end{tabular} &
  \begin{tabular}[c]{@{}l@{}}Optimal op schedules to\\ minimize mem. footprint\end{tabular} &
  \begin{tabular}[c]{@{}l@{}}No memory allocation \\No tensor replacement\end{tabular} \\ \midrule
\begin{tabular}[c]{@{}l@{}}TelaMalloc \\ \cite{telamalloc}\end{tabular} &
  \begin{tabular}[c]{@{}l@{}}ILP/CP-based opt. for \\ memory allocation \\ under memory budget\end{tabular} &
  \begin{tabular}[c]{@{}l@{}}No operator scheduling\\  No tensor replacement \end{tabular} \\ \midrule
MODeL\cite{model} &
  \begin{tabular}[c]{@{}l@{}}ILP-based opt. for \\ scheduling and allocation \\ to reduce memory usage\end{tabular} &
  \begin{tabular}[c]{@{}l@{}}No allocation and replacement\\  solution if spilling required
  \\ Does not scale to complex DNNs
  \end{tabular} 
  \\ \midrule
  \begin{tabular}[c]{@{}l@{}} \name \\ (our work) \end{tabular} &
  \begin{tabular}[c]{@{}l@{}}ILP-based opt. combining \\ ordering, allocation \\ and replacement\end{tabular} &
  \begin{tabular}[c]{@{}l@{}}Additional Divide-and-Conquer \\ technique for scalability 
  \end{tabular} \\ \bottomrule
\end{tabular}

\end{table}



\textbf{Operator Scheduling}
determines the {\em memory footprint}, i.e., the memory used at any time, for executing the DNN on the target hardware, as the operator order determines which tensors are alive at any time. 
Works such as Serenity~\cite{serenity} and HMCOS~\cite{hmcos} can generate an optimal operator schedule with {\em minimal peak memory footprint} (MPMF). 
The actual memory requirement is often larger than the peak memory footprint due to memory fragmentation (as pointed out in their evaluation in the papers), e.g., even with enough total memory to fit all tensors, the there may not be contiguous space needed to place individual tensors.
This results in additional off-chip data accesses to move tensors back and forth from host memory. 
Thus, these works do not co-optimize for memory allocation and tensor replacement.

\textbf{Memory Allocation}
determines the placement of the tensors in the accelerator's memory.
Mainstream DNN compiler frameworks~\cite{pytorch, tensorflow} organize hardware memory in a way similar to a traditional dynamic memory allocator, which cannot fully utilize the memory space and frequently leads to memory fragmentation.
This requires expensive off-chip data accesses to adjust the tensor locations to ``repack" the memory.
A recent work, TelaMalloc~\cite{telamalloc}, provides an optimal memory allocation by encoding the problem using Integer-Linear Programming (ILP) or Constraint-Programming (CP) that can find a valid layout to place all the tensors under a given memory budget.
However, this requires a fixed operator schedule as input, and does not provide a solution, such as tensor replacement, if no feasible allocation scheme can be found.


\textbf{Tensor Replacement} decides which tensor to evict when space is needed. This replacement directly determines the amount of non-compulsory off-chip data accesses, and also influences subsequent memory allocation. Non-compulsory accesses are accesses over and beyond the compulsory accesses which include the initial transfer of a tensor from host processor memory to accelerator memory and the final transfer of the result tensors from accelerator memory to host processor memory.
Common replacement heuristics such as ``Least-Recently Used'' generally cannot provide an optimal solution.
Belady's algorithm is an optimal replacement policy in traditional cache replacement where all data have the same size~\cite{belady}, but it is no longer optimal when tensors have different shapes and sizes~\cite{belady-not-opt}. 

A recent work, MODeL~\cite{model}, proposed an ILP-based approach to jointly determine DNN operator scheduling and memory allocation to minimize the memory usage.
However, its ILP formulation does not support tensor replacement, and thus it does not
minimize off-chip data accesses when the memory budget is insufficient to store all needed tensors.

\name is the first framework to include the aforementioned three dimensions in the optimization of off-chip accesses in mapping DNN applications to accelerators. 
Table~\ref{tab:my-table} summarizes a high-level comparison with related work.
\name can be easily used to generate solutions to sub-problems arising from these three dimensions, e.g., finding operator schedules with minimum peak memory footprint, or finding optimal memory allocation and tensor replacement given fixed schedules.
Further, for complex DNNs with size beyond the scope of existing solutions, \name also provides a divide-and-conquer heuristic that generates high quality solutions in reasonable time.



\section{\name Framework}

\subsection{Target DNN and Accelerator}
In the mapping of DNN computation graphs to accelerators, we treat an operator as a single unit. 
Thus, optimizing techniques used in mapping a single operator, such as operator tiling, are not the focus of this paper, and \name could work orthogonally with such techniques.
The computation of the DNN is statically determined, which applies to the majority of the state-of-the-art DNNs.
The target accelerator has a software-controlled scratchpad memory, where the memory address and access are fully controlled by the programmer/compiler.
The memory allocation can manage either a single scratchpad for the accelerator (or individual scratchpads of the parallel processing elements in the accelerator).
Once a tensor is allocated in memory, its position cannot be changed unless it is evicted and re-allocated.
This memory model applies to accelerators used in real-world edge and server workloads~\cite{telamalloc}.
As in existing work, we use a discrete timestep to model time, in which each timestep represents the execution of one operator in the DNN~\cite{telamalloc, model}.




\subsection{Encoding of Operator Scheduling and Tensor Replacement}
Similar to the ILP encoding proposed by MODeL~\cite{model}, \name models the execution of the DNN by defining the actions of the tensors which are associated with the DNN operators. 
We use the following four sets of binary variables to capture the actions on the tensors, including creation, preservation, spilling and retrieval of the tensors.
Compared with MODeL, we use two additional sets of variables for spills and retrieval to capture the tensor replacement.

\begin{itemize}[leftmargin=*]
    \item $C_{a, \:t} \in \{0, 1\}$: whether tensor $a$ is created at timestep $t$
    \item $P_{a, \:t} \in \{0, 1\}$: whether tensor $a$ is preserved at timestep $t$, i.e., tensor is kept resident in accelerator memory
    \item $S_{a, \:t} \in \{0, 1\}$: whether tensor $a$ is spilled (also deallocated) from accelerator memory to host memory at timestep $t$
    \item $R_{a, \:t} \in \{0, 1\}$: whether tensor $a$ is retrieved (i.e., read) from host memory back to accelerator at timestep $t$
\end{itemize}

Note that spilling and retrieval only apply to live tensors, and dead tensors are deallocated immediately after execution (modeled by setting their preservation variable to zero).
For the example in Fig.~\ref{fig:overall-prob} (right side), tensor $a_{2}$ is created by the operator executed at $t=1$, spilled at $t=2$ to make room for $a_{3}$, retrieved at $t=3$ to be consumed by node 3, and not used again.
This is captured by 
$C_{a_{2}, \: 1} = 1, S_{a_{2}, \: 2} = 1, R_{a_{2}, \: 3} = 1$, and all other variables for $a_{2}$ are 0.

These variables are used to define constraints that enforce correctness of the operator schedules, and are used in the objective function of our ILP formulation.
In the following, $A$ represents the set of tensors, and $T$ the set of timesteps.

The following set of constraints ensure the correct relation of the four sets of variables for the same tensor.
Eq.~\ref{eq:sch-c-single-act} enforces that there is only one action for each tensor at any timestep. 
Eq.~\ref{eq:sch-c-init-preserve} enforces that a tensor cannot be preserved unless it is resident in memory previously, i.e., it is created, preserved or retrieved at the previous timestep.
Eq.~\ref{eq:sch-c-init-spill} enforces that a tensor cannot be spilled unless it is resident at the previous timestep, and Eq.~\ref{eq:sch-c-init-retrieval} enforces that a tensor cannot be retrieved unless it has been spilled, i.e., the host memory has a copy of the tensor.

\small
\begin{align}
    \label{eq:sch-c-single-act}
    &\forall t \in T, \; \forall a \in A  &&\:C_{a, \:t} + P_{a, \:t} + S_{a, \:t} + R_{a, \:t} \leq 1\\
    \label{eq:sch-c-init-preserve}
    &\forall t \in T, \; \forall a \in A  &&\:P_{a, \:t} \leq C_{a, \:t-1} + P_{a, \:t-1} + R_{a, \:t-1}\\
    \label{eq:sch-c-init-spill}
    &\forall t \in T, \; \forall a \in A  &&\:S_{a, \:t} \leq C_{a, \:t-1} + P_{a, \:t-1}\\
    \label{eq:sch-c-init-retrieval}
    &\forall t \in T, \; \forall a \in A  &&\:R_{a, \:t} \leq \textstyle\sum_{k=0}^{t} S_{a, \:k}
\end{align}
\normalsize


The following set of constraints enforce the relationship between the input and output tensors of an operator. Here, an operator is referred to using a tensor it produces, thus $in(a)$ refers to the input tensors of the operator that produces tensor $a$.
Eq.~\ref{eq:sch-c-data-dependency} enforces data dependency in that all the input tensors of the operator should be resident, i.e., preserved or retrieved, before the execution.
Eq.~\ref{eq:sch-c-create-for-all-output} enforces that all ``sibling" tensors of a tensor, which are created by the same operator, are created at the same timestep.
Eq.~\ref{eq:sch-c-single-create} enforces that there is only one creation, i.e, no re-materialization of tensors. Eq.~\ref{eq:sch-c-single-spill} enforces only one spill for each tensor. (Once a tensor is spilled, no further spilling is needed since host memory already has a copy.)

\small
\begin{align}
    \label{eq:sch-c-data-dependency}
    &\forall t \in T, \; \forall a \in A &&\:\forall b \in in(a), \:C_{a, \:t} \leq P_{b, \:t} + R_{b, \:t}\\
    \label{eq:sch-c-create-for-all-output}
    &\forall t \in T, \; \forall a \in A &&\:\forall b \in sib(a), \:C(a, t) = C(b, t)\\
    \label{eq:sch-c-single-create}
    &\forall a \in A &&\:\textstyle\sum_{\: t \in T} C_{a, \:t} = 1 \\
    \label{eq:sch-c-single-spill}
    &\forall a \in A &&\:\textstyle\sum_{\: t \in T} S_{a, \:t} \leq 1 
\end{align}
\normalsize

\subsection{Encoding of Memory Allocation}
In contrast to MODeL which does not support tensor replacement and thus has a single allocation for tensors over all time, \name encodes time information in the memory allocation formulation. This provides for modeling of tensor location in memory for any given timestep 
using the following variables.

\begin{itemize}[leftmargin=*]
    \item $L_{a, \:t} \in [0, M_{B}]$: an integer value of the base address of tensor $a$ in the memory at timestep $t$, $M_{B}$ is the memory budget
    \item $V_{a, \:t} \in \{0, 1\}$: whether tensor $a$ is persistent between $t-1$ and $t$, i.e., tensor $a$'s address stays the same between $t-1$ and $t$
    \item $u_{a, \:b, \:t} \in \{0, 1\}$: whether tensors $a$ and $b$ are both in accelerator memory at $t$ and $a$'s base address is larger than $b$'s
    \item $d_{a, \:b, \:t} \in \{0, 1\}$: whether tensors $a$ and $b$ are both in accelerator memory at $t$ and $a$'s base address is smaller than $b$'s
\end{itemize}

In the memory allocation shown in Fig.~\ref{fig:overall-prob} (right side), tensor $a_{1}$ produced by node 1 has base address $L(a_{1}, \:1)$ at $t=1$, and since $a_{1}$ stays in memory and is consumed by node 2 at the next timestep, $V(a_{1}, \:2)=1$. 
Tensor $a_{2}$, produced by the same node 1, resides within memory with $a_{1}$ at $t=1$, and $a_{2}$ is allocated physically below $a_{1}$ in memory, thus $u(a_{1}, \:a_{2}, \:1) = 1$ and $d(a_{1}, \:a_{2}, \:1) = 0$.
The following constraints on these variables enforce a valid memory allocation at each timestep.

In Eq.~\ref{eq:mem-alloc-c-no-overflow}, for every tensor $a$ at any given time, the allocated base address plus the size of $a$ is within the memory budget.
\begin{equation}
\small
\label{eq:mem-alloc-c-no-overflow}
\begin{aligned}
    &\forall t \in T, \; \forall a \in A & &\:L_{a, \:t} \: + \: Size(a) \leq M_{B}
\end{aligned}
\end{equation}

Eq.~\ref{eq:mem-alloc-c-no-overlap} enforces that two resident tensors in memory do not overlap at any time. 
Here, at any time $t$ and for any tensor pair $a$ and $b$, if they both reside in memory, then $u_{a, \:b, \:t}$ and $d_{a, \:b, \:t}$ are two complementary binary values. $M$ is a large positive value.
If $a$ is placed above $b$, i.e., $u_{a, \:b, \:t}=1$ and $d_{a, \:b, \:t}=0$, Eq.\ref{eq:mem-alloc-c-no-overlap} will force $L_{b, \:t} + Size(b) \leq L_{a, \:t}$ to avoid overlapping of two tensors.
The other constraint $L_{a, \:t} + Size(a) \leq M$ will become redundant.
\begin{equation}
\small
\label{eq:mem-alloc-c-no-overlap}\
\begin{aligned}
    \forall t \in T, \forall (a, b) \in A^{2} \;\;\;
        u_{a,\: b,\: t} + d_{a,\: b,\: t} &\leq 1 \\
        u_{a,\: b,\: t} + d_{a,\: b,\: t} &\geq res_{a,\: t} + res_{b,\: t} - 1 \\
        L_{a, t} + Size(a) - L_{b,\: t} &\leq  M \times (1 - d_{a,\: b,\: t}) \\
        L_{b, t} + Size(b) - L_{a,\: t} & \leq  M \times (1 - u_{a,\: b,\: t}) \\
        \text{where  } \; res_{a,\: t} &= C_{a,\: t} + P_{a,\: t} + R_{a,\: t} \\
        \text{and  } \; res_{b,\: t} &= C_{b,\: t} + P_{b,\: t} + R_{b,\: t}
\end{aligned}
\end{equation}

If a tensor is either created, preserved, or retrieved at the last timestep, and is preserved at the current timestep, its memory address should stay the same, which is enforced by Eq.~\ref{eq:mem-alloc-c-persistent}. 
This is a common ILP encoding of this implication logic using the variable $V_{a, \:t}$ (defined above) for each tensor $a$ at time $t$.
\begin{equation}
\small
\label{eq:mem-alloc-c-persistent}\
\begin{aligned}
    \forall t \in T, \forall a \in A \;\;\;
        L_{a, \: t-1} & \leq L_{a, \: t} + M \times (1 - V_{a, \: t}) \\
        L_{a, \: t} & \leq L_{a, \: t-1} + M \times (1 - V_{a, \: t}) \\
        V_{a, \: t} & \geq (C_{a, \: t-1} + P_{a, \: t-1} + R_{a, \: t-1}) + P_{a, \: t} - 1 \\ 
        V_{a, \: t} & \leq (C_{a, \: t-1} + P_{a, \: t-1} + R_{a, \: t-1}) \\
        V_{a, \: t} & \leq P_{a, \: t}
\end{aligned}
\end{equation}

\subsection{Objective Function to Minimize Off-chip Data Access}
Since \name does not consider decomposition of a single operator (e.g., through operator tiling), the minimum memory budget needed is the maximum over all operators of the sum of the sizes of the input and output tensors for an operator.
Given any memory budget greater or equal than this value, \name can find a solution that minimizes off-chip data accesses. 
Since the compulsory data movement of the input and final output tensors of the DNN is fixed, \name minimizes the total off-chip data accesses by setting the objective function to minimize the \emph{non-compulsory} off-chip data accesses from spilling and retrieval of the tensors, as shown in Eq.~\ref{ilp-obj-min-cla}.
The result considers the operator scheduling (determined by the creation variable of its output tensor), the spilling and retrieval of the tensors, and the memory allocation address of the tensors at each timestep.
\begin{equation}
\label{ilp-obj-min-cla}
\small
\begin{aligned}
    {\argmin}_{\footnotesize \:C,\: P,\: S,\: R,\: L}& & &
    \textstyle\sum_{\: t \: \in \: T} \textstyle\sum_{\: a \: \in \: A} (S_{a, t} + R_{a, t}) \times Size(a) \\ 
    \text{subject to}& & &(\ref{eq:sch-c-single-act}), (\ref{eq:sch-c-init-preserve}), 
    (\ref{eq:sch-c-init-spill}), (\ref{eq:sch-c-init-retrieval}), 
    (\ref{eq:sch-c-data-dependency}), (\ref{eq:sch-c-create-for-all-output}), (\ref{eq:sch-c-single-create}), (\ref{eq:sch-c-single-spill}),
    \\ & & &
    (\ref{eq:mem-alloc-c-no-overflow}), (\ref{eq:mem-alloc-c-no-overlap}), (\ref{eq:mem-alloc-c-persistent})
\end{aligned}
\end{equation}

\subsection{Versatile Objectives}
\name can be adapted to support different use cases.

\subsubsection{Minimizing Peak Memory Footprint}
\label{subsec:ext-min-mem-fp}
\name can generate an operator schedule with an exact minimum peak memory footprint. 
This can be achieved by disabling tensor spilling and retrieval, which is enforced by Eq.~\ref{eq:sch-c-no-cla}.
In addition, a new integer variable $M_{peak}$ is introduced and Eq.~\ref{eq:sch-c-cap-memfpt} ensures that for every timestep, the total size of resident tensors in memory is no larger than $M_{peak}$.
For this purpose, memory allocation is not considered, thus the objective function is set to minimize $M_{peak}$ given only the node ordering constraints.

\small
\begin{align}
    \label{eq:sch-c-no-cla}
    \forall t \in T, \forall a \in A & & &\:S_{a,\: t} = 0, \; R_{a,\: t} = 0\\
    \label{eq:sch-c-cap-memfpt}
    \forall t \in T & & &\:\textstyle\sum_{\: a \: \in \: A} (C_{a,\: t} + P_{a,\: t}) \times Size(a) \leq M_{peak} \\
    {\argmin}_{\footnotesize \: C,\: P}& & &M_{peak} \\ 
    \text{subject to}& & &(\ref{eq:sch-c-single-act}), (\ref{eq:sch-c-init-preserve}), 
    (\ref{eq:sch-c-data-dependency}), (\ref{eq:sch-c-create-for-all-output}), (\ref{eq:sch-c-single-create}),
    (\ref{eq:sch-c-no-cla}), (\ref{eq:sch-c-cap-memfpt}) \nonumber
\end{align}
\normalsize



\subsubsection{Optimizing Memory Allocation and Tensor Replacement for Fixed Schedules}
\label{subsec:ext-fix-order}
Given a fixed schedule, \name  can generate the optimal memory allocation and tensor replacement that minimizes off-chip data accesses under a given memory budget with objective function as in Eq.~\ref{eq:ilp-obj-fix-order}.
Here, Eq.~\ref{eq:sch-c-fix-order} enforces the node ordering by fixing the create variables of the tensors (converted to linear constraints for the ILP encoding). 


\small
\begin{align}
    \label{eq:sch-c-fix-order}
    \forall t \in T, \forall a \in A & & &\: \text{if } order(a) = t, C_{a,\: t} = 1, \text{ else } C_{a,\: t} = 0 \\
    \label{eq:ilp-obj-fix-order}
    {\argmin}_{\: C,\: P,\: S,\: R,\: L}
    & & &\textstyle\sum_{\: t \: \in \: T} \textstyle\sum_{\: a \: \in \: A} (S_{a, t} + R_{a, t}) \times Size(a) \\ 
    \text{subject to}& & &(\ref{eq:sch-c-single-act}), (\ref{eq:sch-c-init-preserve}), (\ref{eq:sch-c-init-spill}), (\ref{eq:sch-c-init-retrieval}), (\ref{eq:sch-c-data-dependency}), (\ref{eq:sch-c-create-for-all-output}), (\ref{eq:sch-c-single-create}), (\ref{eq:sch-c-single-spill}), \nonumber \\
    & & &(\ref{eq:mem-alloc-c-no-overflow}), (\ref{eq:mem-alloc-c-no-overlap}), (\ref{eq:mem-alloc-c-persistent}), (\ref{eq:sch-c-fix-order}) \nonumber
\end{align}
\normalsize

\subsection{Complexity Analysis}
Since we need to create variables and constraints for each tensor pair for each timestep for the memory allocation decisions, 
the size of the ILP encoding is $O({|T|\times|A|}^{2})$, 
where $|T|$ equals to the number of operators in the DNN and $|A|$ is the number of tensors.
However, the number of these variables and constraints is lower in practice because not every tensor pair can overlap in time, e.g., in Fig.~\ref{fig:overall-prob}, $a_{0}$ and $a_{6}$ would not overlap, since $a_{0}$ is consumed by both of the predecessor tensors of $a_{6}$ and would be deallocated by running operator 5 (which generates $a_{6}$).
We use a tensor overlap analysis (similar to strategies in MODeL) to reduce the number of variables and constraints based on the minimum/maximum timesteps (or maximum lifetime) of tensors and data dependencies between tensors.
The majority of the human-designed DNNs have a streamlined structure which greatly reduces the number of variables and constraints.
An example of a popular DNN designed by human experts, ResNet~\cite{resnet}, is shown in Fig.~\ref{fig:resnet-graph}, which has very limited parallel branches of a small number of operators.
This leads to small maximum lifetime for the tensors and greatly reduces the number of overlapping lifetime tensor pairs to be considered.

By supporting tensor replacement and memory allocation optimization, the number of variables in \name is $O({|T|\times|A|}^{2})$, compared with MODeL for which it is $O({|T|\times|A|})$. For a concrete example of optimizing the Transformer~\cite{transformer} model in \S\ref{sec:eval}, the number of variables increased from about 2,000 with MODeL's encoding to 7,500 in \name, and the number of constraints increased from 5,000 to 15,000. However, even with this $\approx 3\times$ increase in variables and constraints, \name can generate an optimal solution within reasonable time as shown in \S\ref{sec:eval}.



\section{Scaling to Complex DNNs}


\begin{figure}[t]
\captionsetup{font=small}
\centering
    \begin{subfigure}{0.48\textwidth}
        \centering
        \includegraphics[width=0.9\linewidth]{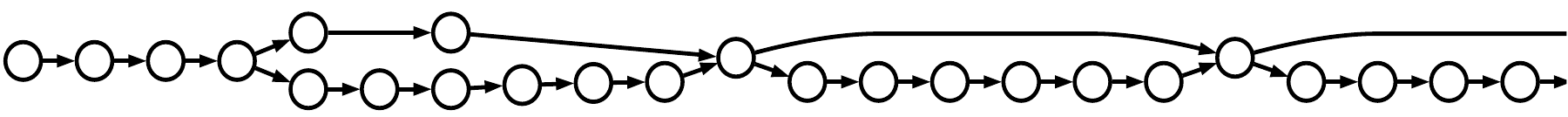}
        \caption{Graph snippet of ResNet~\cite{resnet}, designed by human experts.}
        \label{fig:resnet-graph}
    \end{subfigure}
    \begin{subfigure}{0.48\textwidth}
        \centering
        \includegraphics[width=\linewidth]{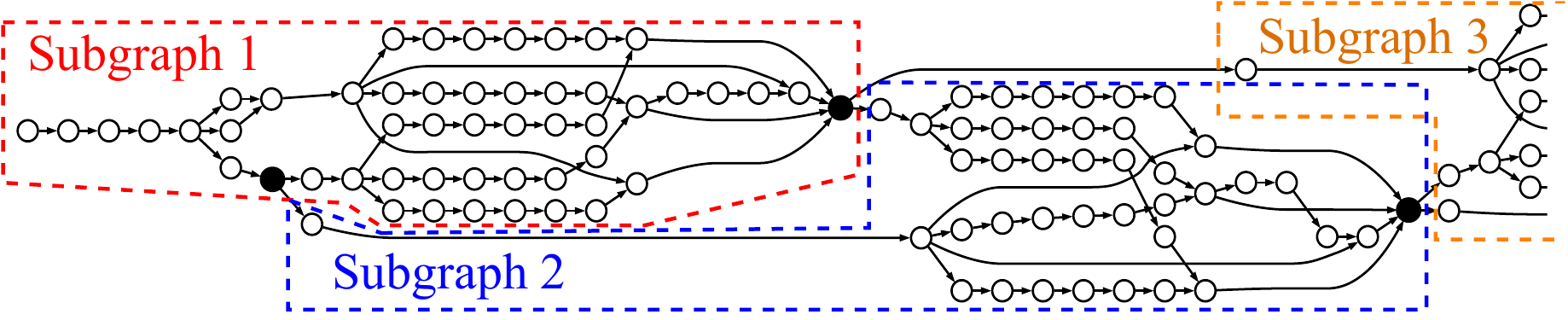}
        \caption{Graph snippet of DARTS~\cite{darts}, designed by NAS.}
        \label{fig:nas-dnc}
    \end{subfigure}
    \caption{Graph structures of DNNs designed by human and by NAS.
    Fig.~\ref{fig:nas-dnc} also illustrates the divide-and-conquer technique to divide the original graph into subgraphs by selecting certain operators as break-nodes (marked as solid circles.)
    }
    \label{fig:model-graphs}
\vspace*{-0.5cm}
\end{figure}
There has been success in developing DNNs using Neural Architecture Search (NAS)~\cite{nas-survey}, which provides automation in exploring competing DNN structures under specific constraints.
Certain state-of-the-art NAS generated DNNs~\cite{nasnet, darts} exhibit a much more complex graph structure than human-designed models. These NAS-generated DNNs have large parallel branches of operators with irregular wiring, as shown in Fig.~\ref{fig:nas-dnc}. This structure poses a practical challenge for the optimization formulation presented thus far and these instances time out even with state-of-the art ILP solvers.

To address this challenge, \name uses a divide-and-conquer based heuristic that can scale up solutions to support these complex NAS models.
We select certain operators in the DNN graph 
as ``break nodes," which divide the DNN graph into several smaller sub-graphs.
The output tensors of these break nodes are treated as the final output tensors of its sub-graph, and as the input tensors for the following connected sub-graphs.
For example, in Fig.~\ref{fig:nas-dnc}, the operators represented as solid circles are selected as ``break nodes," and the original graph can be divided into three connected sub-graphs.
Then \name applies the exact ILP formulation to minimize off-chip data accesses to generate the local optimal solution for each sub-graph, and then combines these into a final global schedule, memory allocation, and tensor replacement of the original DNN.
This combined solution is not guaranteed to be globally optimal as: (i) it does not allow scheduling of operators across different sub-graphs, and (ii) it may incur data accesses for the outputs of the "break nodes" that may not be needed in an optimal solution.


\section{Evaluation}
\label{sec:eval}

\subsection{Experiment Setup}
\name takes the PyTorch implementation of a DNN and converts it to an internal dataflow graph that is then encoded into an ILP instance.
We use Gurobi~\cite{gurobi}, an off-the-shelf general optimization solver, as the ILP solver, and an Apple laptop with M1Pro CPU and 16GB main memory for all experiments.

\subsubsection{DNNs}
We evaluated \name on a wide range of state-of-the-art human-designed DNNs covering different application domains: image classification (ResNet-50~\cite{resnet}, DenseNet~\cite{densenet}, ResNeXt~\cite{resnext}), video classification (R2Plus1D~\cite{r2plus1d}, S3D~\cite{s3d}), semantic segmentation (FCN~\cite{fcn}, L-RASPP~\cite{lraspp}, DeepLabV3~\cite{deeplabv3}), and transformer-based models (Transformer~\cite{transformer} and ViT~\cite{vit}).
In addition, we selected four state-of-the-art DNNs developed by NAS: PNASNet-5~\cite{pnasnet}, AmoebaNet-D~\cite{amoebanet}, NASNet-A~\cite{nasnet} and DARTS~\cite{darts}.
These NAS models all exhibit complex graph structure and irregular wiring between nodes.

\subsubsection{Comparison Schemes}
\label{subsec:eval-benchmarks}
Since \name is the first approach that considers three dimensions in its optimization, there is no tool/technique that is available for a complete comparison. 
Specifically, neither TelaMalloc~\cite{telamalloc} nor MODeL~\cite{model} can support tensor replacement decisions; therefore, they cannot be used for comparison.
Instead, we compare it against approaches used in practice by tools
that handle some variant of the three.
\begin{itemize}[leftmargin=*]
    \item Operator Scheduling: We consider two options for comparison:
   (i) The {\em default} operator scheduling from the PyTorch implementation of the DNNs, and (ii) 
   the operator schedule with {\em minimum  peak memory footprint} (MPMF) of the DNN. 
    \item Memory Allocation: We compare against a linear allocation scheme used in the TensorFlow Lite~\cite{tensorflow-lite-memory-alloc}, which is designed for memory allocation for hardware accelerators.
    \item Tensor Replacement: We use two different tensor replacement policies for comparison: (i) 
    {\em Belady}'s algorithm, which always replaces the tensor to be accessed in the furthest future, and (ii)
    an ILP-based {\em greedy} algorithm which provides a locally optimal decision that generates the least off-chip data accesses each time replacement is needed. 
\end{itemize}

Specifically, we compare \name against four other \emph{schemes}, 
each of which uses the linear memory allocator from TensorFlow Lite, and a distinct combination from the two operator schedule choices ({\em default}, {\em MPMF}) and the two tensor replacement policies ({\em Belady}, {\em greedy}) described above.

For each DNN, we 
consider
three distinct memory budgets, as different memory budgets will 
cause differing tightness of constraints.
(i) The first and tightest memory budget is the minimum memory requirement, denoted $M_{R}$, which is the maximum memory required by a single operator in the DNN.
This is the sum of the sizes of all input and output tensors that must reside in memory during the operation.
(ii) The second budget is the minimum peak memory footprint (MPMF) required for executing the entire DNN, denoted $M_{P}$. 
MPMF is calculated:
(i) by using \name for the human-designed 10 DNNS (\S\ref{subsec:ext-min-mem-fp}), and (ii) by using HMCOS for NAS models. 
While {\name} cannot provide the MPMF for the complex NAS DNNs, HMCOS can provide the MPMF for these DNNs -- HMCOS's inability to provide tensor replacement is not a limitation for the limited purpose of providing the MPMF.
Note that $M_{P}$ is always greater than or equal to $M_{R}$, as it accounts for other live tensors not used by an operator.
(iii) The third, $M_{H}$, is equal to $(M_{R} + M_{P}) / 2$, the average of $M_R$ and $M_P$.

\subsection{Performance on Human-Designed DNNs}
We evaluate \name's performance on ten different human-designed DNNs and compare it against the four schemes 
(\S\ref{subsec:eval-benchmarks}).
While DNN accelerators may have separate scratchpads for {\em activation} and {\em parameter} tensors, some only have a unified scratchpad for both types of tensors.
The activation tensors include the DNN input, output, and the intermediate tensors which are used across different operators, while the parameter tensors are used only for single operators, e.g., the weight tensors of a convolution layer. 
Therefore, we consider two different settings of the tensors: (i) {\em only the activation tensors without the parameter tensors} in the DNN, and (ii) {\em both activation tensors and parameter tensors}.

\subsubsection{Time-to-solution}
\name can generate the optimal solutions to minimize off-chip data access within 1 second, 0.296s on average, for nine of the ten DNNs for all three memory budgets and for both tensor settings. DeepLabV3 takes longer when including both parameter and activations, which takes 6.96s and 17.242s under memory budgets $M_{R}$ and $M_{H}$, respectively (0.12s for $M_{P}$). 
This demonstrates that \name can provide quick time-to-solution for various human-designed models under different memory budgets and tensor settings.

\begin{figure*}[t]
    \centering
    \includegraphics[width=0.95\textwidth]{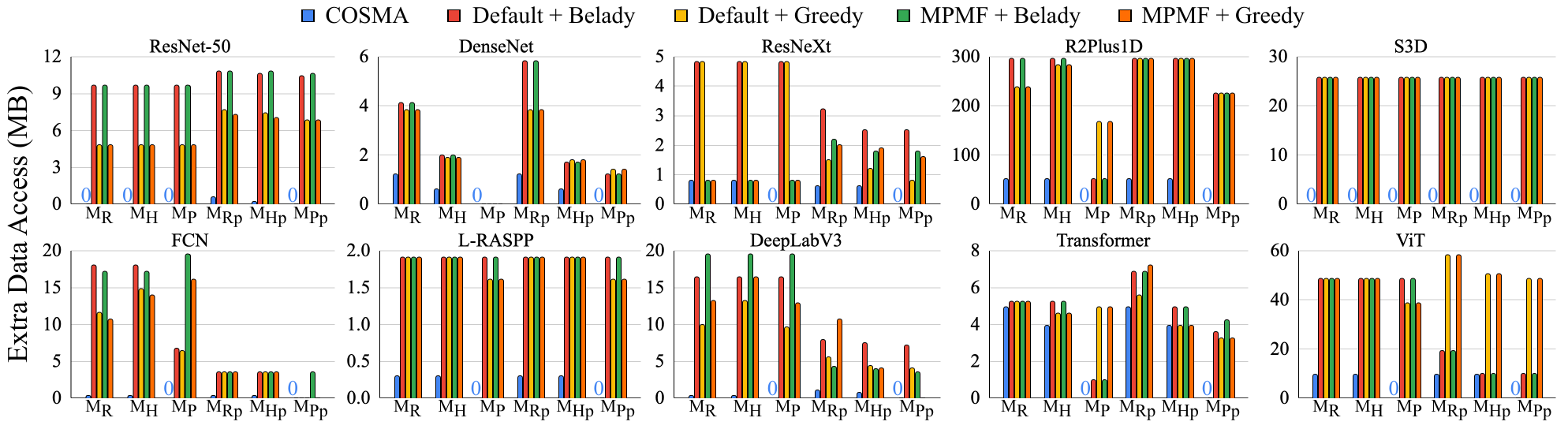}
    \captionsetup{font=small}
    \caption{
        Comparison of off-chip data accesses from tensor spilling and retrieval. For each DNN, $M_{R}, M_{P}, M_{H}$ are the minimum memory required, minimum peak memory footprint, and their average, for activations only. Similarly, $M_{Rp}, M_{Pp}, M_{Hp}$ are the corresponding memory budgets but also take parameter tensors into account. R2Plus1D and S3D take an input tensor of size $(1, 3, 16, 224, 224)$, and Transformer takes two input tensors of size $(10,32,512)$ and $(20,32,512)$. The remaining DNNs have input tensors of size $(1, 3, 224, 224)$. All data are of 8-bit datatype.
    }
    \label{fig:data-mov-comp-reg}
\end{figure*}

\subsubsection{Reduction in Off-Chip Data Accesses}
Fig.~\ref{fig:data-mov-comp-reg} shows the comparison of \name against the four schemes,
for the amount of off-chip data accesses from tensor spilling and retrieval. 
The results demonstrate that given a memory budget of MPMF ($M_{P}$ and $M_{Pp}$), \name can always eliminate the non-compulsory off-chip data accesses due to tensor spilling and retrieval.
This is possible since this memory budget equals the maximum sum of the all live tensors at any timestep.
However, all other schemes incur non-compulsory data accesses except for the model FCN.
In all benchmarks, \name always generates fewer off-chip data accesses compared to other schemes. 
In particular, under the tightest memory limit of $M_{R}$, \name can reduce on average 84\% of non-compulsory data accesses for these ten DNNs under different memory budgets.
Our results also show 
that although operator schedule with MPMF can reduce the off-chip data accesses compared to the default PyTorch operator schedule under the same tensor replacement policy, it still generates significantly more non-compulsory data accesses than \name. 
This shows that optimizing operator scheduling alone is not effective in reducing memory traffic under a tight memory budget.
Further, the results also demonstrate that Belady's algorithm is \emph{not} the optimal tensor replacement policy for mapping DNNs to accelerators, as it generates more off-chip data accesses than even the greedy tensor replacement policy for the same operator schedule and memory allocation.

\subsection{Performance on Complex NAS DNNs}
Next, we evaluate \name on four state-of-the-art DNNs developed by NAS. These contain wide parallel branches of operators and irregular connections, i.e., data dependencies, between the operators.
Without the Divide-and-Conquer (DnC) heuristic enabled, \name cannot generate the optimal solutions for these DNNs within a 24-hour time limit.
In addition to the DnC heuristic, we consider another heuristic which fixes the operator schedule and enables \name to find the globally optimal memory allocation and tensor replacement.
Fixing the schedule can greatly reduce the complexity of the problem, and thus \name can solve for the rest in reasonable time. We refer to these two heuristics as \name DnC and \name FS (Default)/\name FS (MPMF) depending on whether the {\em default Pytorch schedule} or the {\em MPMF schedule} generated by HMCOS~\cite{hmcos} is used. 
Since HMCOS does not support optimization including parameter tensors, we only consider activation tensors in the evaluation for NAS DNNs.

\subsubsection{Time-to-Solution}
Table~\ref{tab:tts-nas-model} shows the solving time of the different heuristics on the four DNNs: (i) \name FS (Default) (ii) \name FS (MPMF), and (iii) \name DnC.
\name FS (Default) and \name FS (MPMF) can generate the optimal memory allocation and tensor replacement for these four models under 30 seconds for most cases, with marginal variance between different memory budgets. The exception is AmoebaNet-D, which takes 84.4 seconds for \name FS (MPMF) under the tightest memory budget $M_{R}$. 
After manually selecting operators as break nodes in the DNN graph, \name DnC can generate solutions within 2 minutes, and the time-to-solution is more sensitive to different memory budgets. (Future work will consider automating the selection of break nodes.) 

\begin{table}[t]
\captionsetup{font=small}
\caption{Time-to-solution (in seconds) for \name for the complex NAS DNNs. The left two sets of columns are for \name \underline{f}ixed operator \underline{s}chedule (FS) heuristics, 
and the right set of columns are for the \name Divide-and-Conquer (DnC) heuristic. The schedule with minimum peak memory footprint (MPMF) is generated by HMCOS.
}
\setlength\tabcolsep{4pt}
\centering
\label{tab:tts-nas-model}
\begin{tabular}{@{}l|lcc|ccc|ccc@{}}
\toprule
\multicolumn{1}{c|}{\multirow{3}{*}{NAS DNN}} & \multicolumn{3}{c|}{\name FS} & \multicolumn{3}{c|}{\name FS} & \multicolumn{3}{c}{\multirow{2}{*}{\name DnC}} \\
\multicolumn{1}{c|}{} & \multicolumn{3}{c|}{(Default)} & \multicolumn{3}{c|}{(MPMF)} & \multicolumn{3}{c}{}        \\
\multicolumn{1}{c|}{} & $M_R$   & $M_{H}$   & $M_{P}$  & $M_{R}$ & $M_{H}$ & $M_{P}$ & $M_{R}$ & $M_{H}$ & $M_{P}$ \\ \midrule
PNASNet-5             & 12.2    & 12.3      & 12.2     & 24.2    & 22.4    & 23.5    & 65.9    & 63.5    & 58.0    \\
AmoebaNet-D           & 8.1     & 8.3       & 8.1      & 84.4    & 8.2     & 8.5     & 62.1    & 59.9    & 112.8   \\
NASNet-A              & 19.8    & 20.0      & 19.7     & 22.4    & 22.4    & 21.6    & 37.5    & 22.9    & 39.2    \\
DARTS                 & 5.0     & 5.0       & 5.0      & 5.2     & 5.2     & 5.2     & 19.1    & 19.1    & 19.1    \\ \bottomrule
\end{tabular}

\end{table}
\begin{figure}[t]
    \centering
    \includegraphics[width=0.48\textwidth]{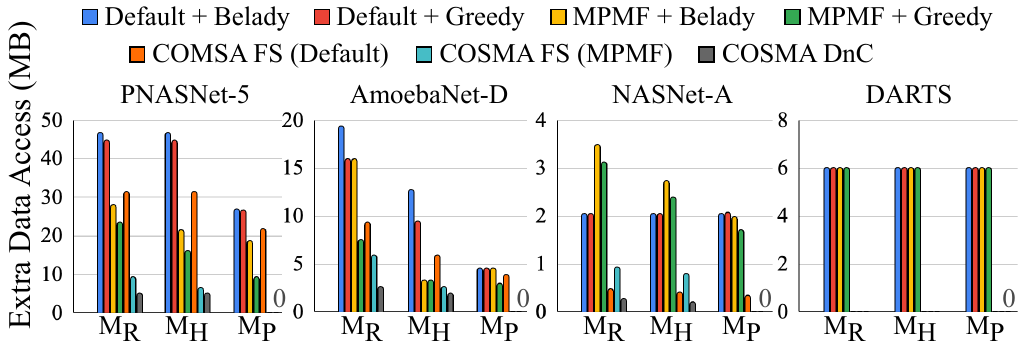}
    \captionsetup{font=small}
    \caption{Non-compulsory off-chip data access volume for different approaches on the NAS DNNs. All DNNs take an input tensor of size $(1, 3, 224, 224)$ with 8-bit datatype.}
    \label{fig:data-mov-comp-nas}
    \vspace*{-0.5cm}
\end{figure}

\subsubsection{Reduction in Data Accesses}
Fig.~\ref{fig:data-mov-comp-nas} summarizes the non-compulsory off-chip data accesses generated by the different approaches -- 
\name DnC, \name FS (Default)/\name FS (MPMF), and the four comparison schemes -- 
on the four NAS DNNs.
Note that although \name DnC may not generate globally optimal solutions, it still generates the least number of off-chip data accesses among all, and eliminates tensor spilling and retrieval with the memory budget $M_{P}$.
Further, given the same fixed operator schedules, the two \name FS heuristics can find the optimal memory allocation and tensor replacement that greatly reduces the off-chip data accesses, in comparison to the memory allocator from TensorFlow Lite and with either Belady's replacement policy or the greedy algorithm.
Further, the results again show that minimizing the peak memory footprint of the operator schedule does not necessarily lead to the least off-chip data accesses, and it even generates more data accesses compared to the default schedules in the case of NASNet-A under tighter memory budgets ($M_{R}$ and $M_{H}$).





\section{Conclusion}
We propose \name, a framework that considers the three dimensions of operator scheduling, memory allocation, and tensor replacement decisions to minimize off-chip data accesses for mapping DNNs to a target accelerator.
We formulate the optimization problem using ILP and provide an encoding of the tensors' memory status and memory location throughout the execution of the DNN on a target accelerator.
We demonstrate that \name can generate optimal solutions in sub-seconds on average, using an off-the-shelf ILP solver, for ten popular human-designed DNNs from various application domains, and reduces off-chip memory accesses,on average, by 84\% in comparison with existing approaches.
We also explore a Divide-and-Conquer heuristic in conjunction with \name to scale up to complex DNNs developed by Neural Architecture Search (NAS). This generates solutions that reduce, on average, 85\% of non-compulsory data accesses for four state-of-the-art complex NAS DNNs, with an average solution time of one minute.


\bibliographystyle{IEEEtranS}
\bibliography{references}

\begin{thebibliography}{10}
\providecommand{\url}[1]{#1}
\csname url@samestyle\endcsname
\providecommand{\newblock}{\relax}
\providecommand{\bibinfo}[2]{#2}
\providecommand{\BIBentrySTDinterwordspacing}{\spaceskip=0pt\relax}
\providecommand{\BIBentryALTinterwordstretchfactor}{4}
\providecommand{\BIBentryALTinterwordspacing}{\spaceskip=\fontdimen2\font plus
\BIBentryALTinterwordstretchfactor\fontdimen3\font minus \fontdimen4\font\relax}
\providecommand{\BIBforeignlanguage}[2]{{%
\expandafter\ifx\csname l@#1\endcsname\relax
\typeout{** WARNING: IEEEtranS.bst: No hyphenation pattern has been}%
\typeout{** loaded for the language `#1'. Using the pattern for}%
\typeout{** the default language instead.}%
\else
\language=\csname l@#1\endcsname
\fi
#2}}
\providecommand{\BIBdecl}{\relax}
\BIBdecl

\bibitem{tensorflow}
M.~Abadi, P.~Barham, J.~Chen, Z.~Chen, A.~Davis, J.~Dean, M.~Devin, S.~Ghemawat, G.~Irving, M.~Isard \emph{et~al.}, ``{TensorFlow: a system for Large-Scale machine learning},'' in \emph{OSDI}, 2016.

\bibitem{serenity}
B.~H. Ahn, J.~Lee, J.~M. Lin, H.-P. Cheng, J.~Hou, and H.~Esmaeilzadeh, ``{Ordering Chaos: Memory-aware Scheduling of Irregularly Wired Neural Networks for Edge Devices},'' \emph{MLSys}, 2020.

\bibitem{belady}
L.~A. Belady, ``A study of replacement algorithms for a virtual-storage computer,'' \emph{IBM Systems Journal}, 1966.

\bibitem{belady-not-opt}
D.~S. Berger, N.~Beckmann, and M.~Harchol-Balter, ``{Practical Bounds on Optimal Caching with Variable Object Sizes},'' \emph{POMACS}, vol.~2, 2018.

\bibitem{deeplabv3}
L.-C. Chen, G.~Papandreou, F.~Schroff, and H.~Adam, ``{Rethinking Atrous Convolution for Semantic Image Segmentation},'' \emph{arXiv preprint arXiv:1706.05587}, 2017.

\bibitem{chen2018tvm}
T.~Chen, T.~Moreau, Z.~Jiang, L.~Zheng, E.~Yan, H.~Shen, M.~Cowan, L.~Wang, Y.~Hu, L.~Ceze \emph{et~al.}, ``{TVM: An automated End-to-End optimizing compiler for deep learning},'' in \emph{OSDI}, 2018.

\bibitem{vit}
A.~Dosovitskiy, L.~Beyer, A.~Kolesnikov, D.~Weissenborn, X.~Zhai, T.~Unterthiner, M.~Dehghani, M.~Minderer, G.~Heigold, S.~Gelly \emph{et~al.}, ``{An Image is Worth 16x16 Words: Transformers for Image Recognition at Scale},'' \emph{arXiv preprint arXiv:2010.11929}, 2020.

\bibitem{tensorflow-lite-memory-alloc}
\BIBentryALTinterwordspacing
Google. Accessed:2023-09-01. [Online]. Available: \url{github.com/tensorflow/tensorflow/blob/master/tensorflow/lite/simple_memory_arena.cc}
\BIBentrySTDinterwordspacing

\bibitem{gurobi}
\BIBentryALTinterwordspacing
L.~Gurobi~Optimization. Accessed: 2023-09-01. [Online]. Available: \url{https://www.gurobi.com/documentation/current/refman/index.html}
\BIBentrySTDinterwordspacing

\bibitem{resnet}
K.~He, X.~Zhang, S.~Ren, and J.~Sun, ``{Deep Residual Learning for Image Recognition},'' in \emph{CVPR}, 2016.

\bibitem{MarkEnergy}
M.~Horowitz, ``{Computing's energy problem (and what we can do about it)},'' in \emph{ISSCC}, 2014.

\bibitem{lraspp}
A.~Howard, M.~Sandler, G.~Chu, L.-C. Chen, B.~Chen, M.~Tan, W.~Wang, Y.~Zhu, R.~Pang, V.~Vasudevan \emph{et~al.}, ``{Searching for MobileNetv3},'' in \emph{ICCV}, 2019, pp. 1314--1324.

\bibitem{densenet}
S.~J{\'e}gou, M.~Drozdzal, D.~Vazquez, A.~Romero, and Y.~Bengio, ``{The One Hundred Layers Tiramisu: Fully Convolutional Densenets for Semantic Segmentation},'' in \emph{CVPR}, 2017, pp. 11--19.

\bibitem{pnasnet}
C.~Liu, B.~Zoph, M.~Neumann, J.~Shlens, W.~Hua, L.-J. Li, L.~Fei-Fei, A.~Yuille, J.~Huang, and K.~Murphy, ``{Progressive Neural Architecture Search},'' in \emph{ECCV}, 2018.

\bibitem{darts}
H.~Liu, K.~Simonyan, and Y.~Yang, ``{DARTS: Differentiable Architecture Search},'' \emph{ICLR}, 2019.

\bibitem{fcn}
J.~Long, E.~Shelhamer, and T.~Darrell, ``{Fully Convolutional Networks for Semantic Segmentation},'' in \emph{CVPR}, 2015, pp. 3431--3440.

\bibitem{telamalloc}
M.~Maas, U.~Beaugnon, A.~Chauhan, and B.~Ilbeyi, ``{TelaMalloc: Efficient On-Chip Memory Allocation for Production Machine Learning Accelerators},'' in \emph{ASPLOS}, 2022.

\bibitem{pytorch}
A.~Paszke, S.~Gross, F.~Massa, A.~Lerer, J.~Bradbury, G.~Chanan, T.~Killeen, Z.~Lin, N.~Gimelshein, L.~Antiga \emph{et~al.}, ``{Pytorch: An imperative style, high-performance deep learning library},'' \emph{NeurIPS}, 2019.

\bibitem{amoebanet}
E.~Real, A.~Aggarwal, Y.~Huang, and Q.~V. Le, ``{Regularized Evolution for Image Classifier Architecture Search},'' in \emph{AAAI}, 2019.

\bibitem{nas-survey}
P.~Ren, Y.~Xiao, X.~Chang, P.-Y. Huang, Z.~Li, X.~Chen, and X.~Wang, ``{A Comprehensive Survey of Neural Architecture Search: Challenges and Solutions},'' \emph{ACM Computing Surveys}, vol.~54, 2021.

\bibitem{model}
B.~Steiner, M.~Elhoushi, J.~Kahn, and J.~Hegarty, ``{MODeL: Memory Optimizations for Deep Learning},'' in \emph{ICML}, 2023.

\bibitem{r2plus1d}
D.~Tran, H.~Wang, L.~Torresani, J.~Ray, Y.~LeCun, and M.~Paluri, ``{A Closer Look at Spatiotemporal Convolutions for Action Recognition},'' in \emph{CVPR}, 2018, pp. 6450--6459.

\bibitem{transformer}
A.~Vaswani, N.~Shazeer, N.~Parmar, J.~Uszkoreit, L.~Jones, A.~N. Gomez, {\L}.~Kaiser, and I.~Polosukhin, ``{Attention is All You Need},'' \emph{NIPS}, vol.~30, 2017.

\bibitem{hmcos}
Z.~Wang, C.~Wan, Y.~Chen, Z.~Lin, H.~Jiang, and L.~Qiao, ``{Hierarchical Memory-constrained Operator Scheduling of Neural Architecture Search Networks},'' in \emph{DAC}, 2022.

\bibitem{resnext}
S.~Xie, R.~Girshick, P.~Doll{\'a}r, Z.~Tu, and K.~He, ``{Aggregated Residual Transformations for Deep Neural Networks},'' in \emph{CVPR}, 2017, pp. 1492--1500.

\bibitem{s3d}
S.~Xie, C.~Sun, J.~Huang, Z.~Tu, and K.~Murphy, ``{Rethinking Spatiotemporal Feature Learning: Speed-accuracy Trade-offs in Video Classification},'' in \emph{ECCV}, 2018, pp. 305--321.

\bibitem{nasnet}
B.~Zoph, V.~Vasudevan, J.~Shlens, and Q.~V. Le, ``{Learning Transferable Architectures for Scalable Image Recognition},'' in \emph{CVPR}, 2018.

\end{thebibliography}

\end{document}